\documentclass[10pt,twocolumn,letterpaper]{article}

\usepackage{cvpr}
\usepackage{times}
\usepackage{epsfig}
\usepackage{graphicx}
\usepackage{amsmath}
\usepackage{amssymb}
\usepackage{algorithm}
\usepackage[noend]{algpseudocode}
\usepackage{arydshln} 
\usepackage{enumitem}
\usepackage{setspace}
\usepackage{url}

\cvprfinalcopy 

\def\cvprPaperID{2592} 

\ifcvprfinal\fi
\begin{document}

\title{Learning to Caption Images through a Lifetime by Asking Questions}

\author{
Kevin Shen$^{1,2}$, \hspace{2pt} Amlan Kar$^{1,2}$, \hspace{2pt} Sanja Fidler$^{1,2,3}$\\
$^{1}$ University of Toronto, $^{2}$ Vector Institute, $^{3}$ NVIDIA \\
{\tt\small \{shenkev, amlan, fidler\}@cs.toronto.edu}
}


\maketitle

\begin{abstract}
In order to bring artificial agents into our lives, we will need to go beyond supervised learning on closed datasets to having the ability to continuously expand knowledge. Inspired by a student learning in a classroom, we present an agent that can continuously learn by posing natural language questions to humans.
Our agent is composed of three interacting modules, one that performs captioning, another that generates questions and a decision maker that learns when to ask questions by implicitly reasoning about the uncertainty of the agent and expertise of the teacher. As compared to current active learning methods which query images for full captions, our agent is able to ask pointed questions to improve the generated captions. The agent trains on the improved captions, expanding its knowledge. We show that our approach achieves better performance using less human supervision than the baselines on the challenging MSCOCO~\cite{lin2014microsoft} dataset.
\end{abstract}

\section{Introduction}

Imagine a child that sees a crocodile for the first time. She may likely ask what the animal is called, or where it can be encountered outside the zoo, but probably does not need to be told that it is green or has four legs, and that its sharp teeth can pose danger. Children (and even adults) learn from teachers in an active way: asking questions about concepts that they are unfamiliar or uncertain about. In doing so, they make learning more efficient -- the child who acquires exactly the information they are missing -- and the teacher who answers the question instead of needing to explain many aspects of a concept in full detail. As A.I. becomes more and more integrated in our everyday lives, be it in the form of personal assistants or household robots \cite{vinyals2015neural, mataric2017socially, simo2015neuroaesthetics}, they too should actively seek out missing information from humans -- by asking questions in the form of natural language which non-experts can understand and answer.

Most existing work on scene understanding tasks such as VQA~\cite{balanced_vqa_v2, teney2017tips, wu2017visual, gupta2017survey} and captioning \cite{lin2014microsoft, rennie2016self, dai2017towards} have focused on a closed world setting, i.e. consuming the knowledge provided by a labeled dataset. On the other hand, the goal of active learning is to be able to continuously update the model by seeking for the relevant data to be additionally labeled by a human \cite{settles2012active}. Most active learning approaches, however, ask the human to provide a full labeling of an example, and the main challenge is in identifying the examples to be labeled, to ensure annotation efficiency. In our work, we go beyond this, by endowing the model with the ability to ask for a particular aspect of a label, and do so in natural language in order to unambiguously identify the missing information.

\paragraph{Lifetime learning} In this paper we define lifetime learning as open world learning over multiple sessions. At the beginning of its lifetime, the agent pretrains in a supervised way. It is assumed that data labels are expensive, and therefore only a small portion of the underlying data distribution is represented by the pretraining data. In lifetime learning, the agent is exposed to data sequentially, one batch per session. Unlabelled data arrives from the world and the agent has the ability to query a teacher for full or partial labels to learn new concepts. However, the agent often has a fixed budget for asking and therefore must train in a self-supervised manner by intelligently deciding what to be labelled.

\begin{figure}[t]
\centering
\includegraphics[width=1.0\linewidth]{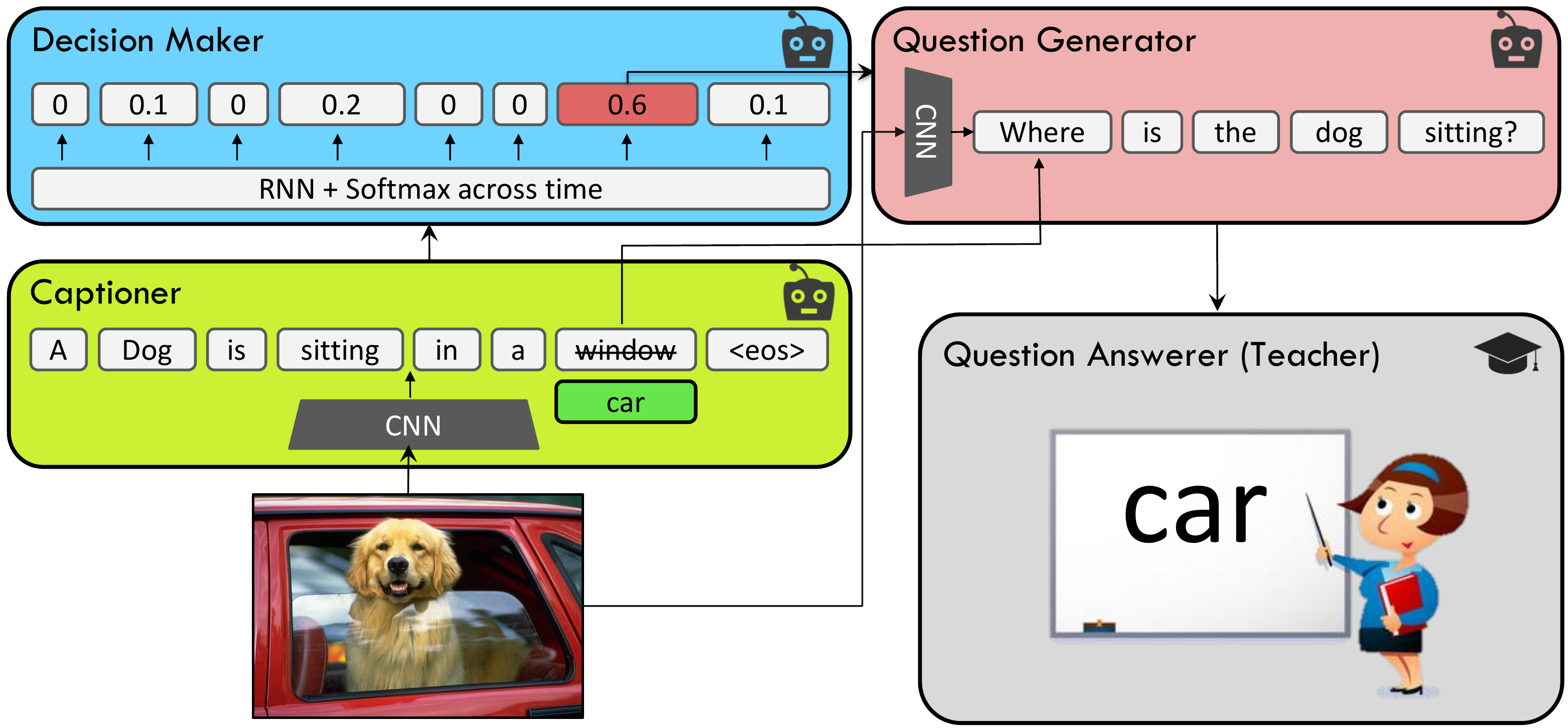}
\caption{Learning to describe images by asking questions. Our model learns in a lifetime learning setting, by actively seeking for missing information. We jointly learn when and what to ask, and learn from the teacher's answers. Our model poses questions in natural language.}
\label{fig:intro}
\end{figure}

We focus on the task of image captioning as a proxy task for scene understanding. In order to describe an image, a model needs to generate words describing the objects, its attributes, actions, and possibly relationships and interactions between objects. This is inherently a multi-task problem. In this paper, our goal is to allow a captioning agent to actively ask questions about the aspects of the image it is uncertain about, in a lifetime learning setting in which examples arrive sequentially and continually. Thus, instead of having humans provide captions for each new image, our agent aims to ask a minimal set of questions for the human to answer, and learn to caption from these answers. 

Our model consists of three modules: a captioning module, a decision making module that learns whether to ask and what to ask about, and a question generation module. At training time when the captioner produces each word, the decision module decides for which concept, if any, to ask about. If the agent decides to ask, the question module produces a question, which the teacher answers. All three modules are implemented as neural networks. They are updated continuously with the data arriving in batches: the captioning module is updated using the captions improved by the answers from the teacher, while the decision module is updated based on the current uncertainty of the captioning module. For efficiency reasons, our teacher to answer questions is a QA bot. At test time the captioning model describes new images without asking questions.\\

\noindent In summary, our contributions are:
\begin{itemize}[]
    \item A new Learning by Asking Questions paradigm in which captioning, question generating, and decision modules interact in order to learn in over a lifetime. The advantage of LBAQ is it improves the efficiency of data collection.
    \item A novel decision maker module, trained with reinforcement learning (RL) that decides whether and what to ask a question about by implicitly reasoning about the uncertainty of the agent and knowledge of the teacher.
\end{itemize}
We showcase our method on the challenging MSCOCO dataset~\cite{lin2014microsoft}. We provide insights into the behavior of our approach, and discuss open challenges ahead. To the best of our knowledge, this is the first time that natural language question asking has been explored in a lifetime learning setting with real-world images. For reproducibility, we have released our code \url{https://github.com/shenkev/Caption-Lifetime-by-Asking-Questions}.
\section{Related Work}
We provide a short overview of active and interactive learning approaches, and outline our main contributions with respect to existing work. 

\paragraph{Active learning.}
The goal of active learning is to intelligently seek labels for unlabelled data from an oracle in order to maximize learning while reducing the annotation cost. An agent predicts which sample, if labelled, will give the most useful learning signal as measured by performance on the test set. Strategies for active learning include uncertainty sampling, query by committee and expected model change~\cite{settles2012active}. Unlike the typical active learning setting where an agent asks the oracle for a full data label (which would be a full caption in our scenario), our method learns to ask pointed questions to retrieve partial labels, \ie missing key words that compose a caption. Our model thus needs to not only learn when to ask, but also what to ask, and how to distill the received answer into a complex multi-task module (captioner). 

\paragraph{Learning by Asking Questions}
is an exciting direction with notable contemporary work. Prior approaches typically differ in task, methodology (are questions natural or templated? how does the agent utilize the feedback?) and environment (synthetic vs real).~\cite{misra2017learning} learns to answer questions by asking questions. Image and the generated question are treated as an unlabelled sample and an oracle provides an answer to form a novel training pair. This simplifies the learning by asking framework by bypassing the challenges of free-form conversation and interpreting the teacher's answer, because QA can be directly used as training data. Our work generalizes over this framework by using question-asking as a support task to the main task, in our case image captioning, which leads to a more general, and significantly more challenging scenario. Furthermore,~\cite{misra2017learning} operates in CLEVR~\cite{johnson2017clevr}, a synthetic environment and questions are limited to programs rather than natural language. 

 \cite{yang2018visual} explores question asking for visual recognition. Given an image, a graph of objects, attributes and relationships is continually updated as the agent asks questions. However, questions are limited to templates, and training is done in synthetic environments with a limited set of objects and relationships. \cite{uehara2018visual} uses questions to explore new object classes for image classification. However,~\cite{uehara2018visual} does not retrain their classifier. Our work differs from~\cite{yang2018visual, uehara2018visual} by proposing a way for the agent to learn in a lifetime setting.
 
In~\cite{weston16}, the agent learns whether to ask questions to the teacher to efficiently solve dialogue tasks. The student's goal is to maximize the accuracy of answering the teacher's questions while reducing the cost (to the teacher) of asking for hints. We extend this line of thinking by letting the agent learn what to ask about in addition to whether to ask.

\paragraph{Vision and Language.} 
Our work tackles captioning~\cite{xu2015show, rennie2016self, dai2017towards}, visual question answering (VQA)~\cite{teney2017tips, gupta2017survey, kim2018progressive}, and visual question generation (VQG)~\cite{li2018visual, mostafazadeh2016generating}.
However, most of these works have focused on a closed dataset setting. Our main goal here is not in designing a novel architecture for each module (captioning, VQG, VQA), but rather focusing on the interaction of the modules and the teacher in order to learn in a continual, active setting.  Related to us is~\cite{ling2017teaching}, where a teacher observes the captioning agent in a continual setting, and gives natural language feedback when errors occur. The agent then learns to improve based on this signal. In our work, the agent is the one seeking advice, thus making 
the teaching process more efficient. 
\section{Our Approach}

Our goal is to train an image captioning model in the active learning setting with minimal human supervision. We approach the problem by endowing the agent with the ability to ask questions, and learn from the teacher's answers. However, question asking is only a tool for retrieving information 
during training; at test time, the captioner operates without needing to ask questions. We first provide an intuitive overview of our interactive training procedure, describing the lifetime learning setting, namely how the agent learns from data arriving in a sequence of batches. Next, we provide details of how the agent queries for, and learns from, answers and feedback from the teacher. Finally, we describe the implementation of our agent's modules.

\begin{figure*}[t]
\centering
\includegraphics[width=1.0\linewidth]{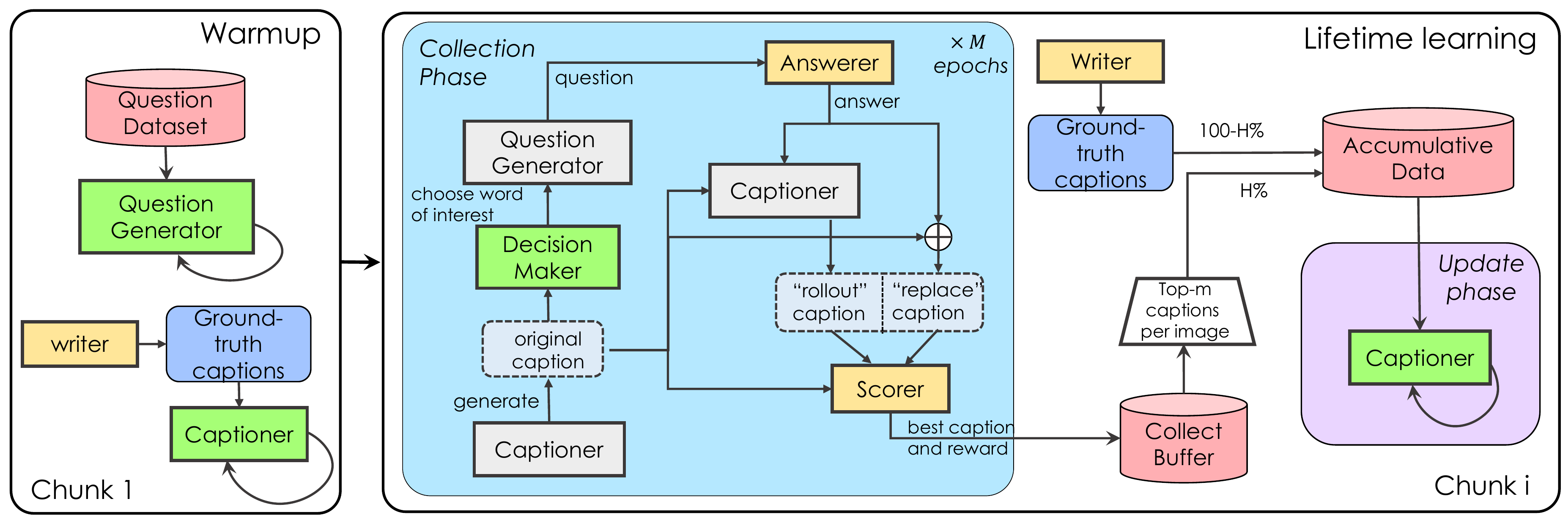}
\caption{Modules being updated (green), modules being held fixed (grey), teacher (yellow). Writer is a teacher that produces full GT captions. The captioner begins by warming up on the first chunk containing all GT captions (left panel). Learning by asking questions (right panel) occurs in two phases: collection and update. In the collection phase, the captioner generates a caption, the decision maker choose when to ask a question, the question generator generates a question and the teacher provides an answer. The answer is used to create two new captions. Captions are collected and used to train the captioner in the update phase.}
\label{fig:model}
\end{figure*}

\subsection{Lifetime learning}
We imagine a lifetime learning setting where data arrives in chunks. This is analogous to a student who learns over multiple classes in a semester. The first chunk $D_w$ has complete ground truth (GT), \ie human written captions. We refer to it as the warmup chunk. The agent learns from the remaining $K$ unlabelled chunks $D_u=[D_{u1}, D_{u2}, \dots, D_{uK}]$ with partial supervision from the teacher. We first train the question generator and pretrain the captioner on the warmup chunk. For each $K^{th}$ unlabelled chunk, the agent iterates between two phases: querying the teacher, and learning from the collected information.

In the \textbf{(caption) collection phase}, the agent interacts with the teacher using two modules: a decision maker, and a question generator. The agent looks at each image in an unlabelled chunk, attempts to caption, and decides whether to replace words with answers obtained by asking questions. The agent collects the improved captions and uses them to train the captioner in the \textbf{(captioner) update phase}. In the collection phase, the feedback from the teacher is also used to train the decision maker to make better decisions about whether and what to ask.
This process is illustrated in Figure~\ref{fig:model}, and summarized in Algorithm~\ref{algo:lifelonglearn}. 

\subsection{Notation\label{sec:modules}}
Let $\mathbf{w} = (w_1,w_2,\dots,$ $w_L)$ denote a caption of length $L$, and $I$ an image. The {\bf captioning module} $C( \mathbf{w} | I)$ computes a probability distribution over the words in a sentence, \ie $p_{\theta_C}(\mathbf{w} | I)$. We further compute $\mathbf{c} = (c_1, c_2, \dots, c_L)$, denoting an array of contexts computed by the captioner (details in Sec~\ref{subsec:modules}). The context helps the decision maker decide what concepts to ask about, and the question generator to ask relevant questions. Let the context used by the decision maker and question generator be called $c^{DM}$ and $c^q$, respectively.
The {\bf decision module} $DM(t | \mathbf{c})$ computes a multinomial distribution $p_{\theta_{DM}}(t| c^{DM})$ indicating the probability of a word position $t$ in the caption at which the question should be asked. We allow $t$ to index a special \texttt{<eos>} position representing the case where no question should be asked. The {\bf question generation module} $Q(\mathbf{q} | I, c^q_t)$ computes the probability distribution $p_{\theta_q}(\mathbf{q} | I, c^q_t)$ over a question $\mathbf{q}$. 
The details about the modules are presented in Sec~\ref{subsec:modules}. 
\subsection{Caption collection phase}
In the collection phase, the agent attempts to improve captions generated from its own policy by querying the teacher. For each round, the agent makes multiple passes over a chunk. Given an image, the agent generates a caption, and the decision maker decides whether and when (at which word) to ask a question to the teacher. The teacher answers the question, which the agent uses to create a new caption (details in Section~\ref{sec:interactteacher}). The teacher scores both new and old captions and the agent stores the captions in a buffer $D_c$. At the same time, the agent uses the scores from the teacher to make online updates to the decision maker to pick better time steps for asking questions (Section~\ref{sec:traindecisionmaker}). 

The collected captions will be used in the update phase by the agent to distill the teacher's knowledge back into the captioner. However, the agent could encounter difficult images that cannot be improved by asking questions. Empirically we find the agent cannot improve on images containing objects in unusual settings, or if the caption generated from the captioner's policy is missing multiple key concepts. Therefore, we allow the agent to ``give up" if the improved caption is bad, and the teacher writes a new caption. This is analogous to a student asking for a full explanation from the teacher after class if he did not understand a concept. For every image, the agent considers the top $m$ captions from the buffer $D_c$ for training. It keeps the top $H\%$ of images-caption tuples based on the average caption reward over $m$ captions. For the other 100-$H\%$ images, the agent ``gives up" and is given $m$ GT captions. In practice, we choose $m=2$ out of the 5 MSCOCO captions. The \texttt{KeepBestAndGiveUp} subroutine in Algorithm~\ref{algo:lifelonglearn} summarizes how the agent selects training data for the captioner.

\subsubsection{Interacting with the Teacher Details \label{sec:interactteacher}}
Given an image, the captioner produces the complete initial caption $\mathbf{w}^0$ and context $\mathbf{c}^0$ by a greedy rollout from $p_{\theta_C}(. | I)$. The decision module then makes a decision by sampling from $p_{\theta_{DM}}(.|c^{DM})$. 
Words other than nouns, verbs, and adjectives are masked out. 
Let $w_t$ be the word for which the decision module decides to ask a question.  The question generator produces a question and the agent receives an answer $a$.  The agent then replaces word $w_t$ in $\mathbf{w}^0$ with $a$ and predicts a new caption $\mathbf{w}^1_{ro} = (w_1 \dots w_{t-1}, a, w'_{t+1}, \dots w'_L)$, by rolling out the rest of the caption from position $t$ using the previous hidden state $h_{t-1}$ of the captioner and $a$. If the teacher's answer is a rare word for the agent, the agent may diverge from any sensible trajectory. For this reason, we give the agent the option of doing a one-word-replace of the expert's answer, \ie $\mathbf{w}^1_{re} = (w_1 \dots w_{t-1}, a, w_{t+1}, \dots w_L)$. 

Finally the teacher scores both the original and the two improved captions, by giving each a numeric reward $r$. The process can be repeated by asking a second question and replacing another word at step $t' > t$. In general, the agent can ask up to $N$ questions for a single caption. In practice, we observe $N=1$ to work  best in our experiments. We keep $N$ in the following for the generality of exposition. The interaction process is summarized in Algorithm~\ref{algo:capbyask}.

\begin{algorithm}[t!]
\caption{Lifetime learning}
\begin{algorithmic}[1]
{\footnotesize
\Procedure{LIFETIME($D_w$, $D_u$)}{}
    \State \textbf{train:} $C$, $Q$, $V$ \Comment{train captioner, question generator, QA-bot}
    \State \textbf{initialize:} DM \Comment{initialize decision maker}
    
    \State $D \gets D_w$
    \State $D_u = [D_{u1}, D_{u2}, \dots D_{uK}]$
    \For{$D_{uk}$ in $D_u$} \Comment{begin lifetime learning}
        \State $D_c \gets [\;]$ \Comment{collection phase}
        \For{$\text{epoch}=1$ to Number of Passes over Chunk}
            \For{$I$ in $D_{uk}$}
                \State $\mathbf{w}, r \gets \texttt{SeekTeacher}(I)$
                \State $\mathbf{w^*}, r^* \gets \texttt{SeekTeacher}(I, \text{ greedy=True})$
                \State $D_c \mathrel{+}= (\mathbf{w}, r, \mathbf{w^*}, r^*)$
                \Comment{collect caps. and rewards}
                
                \State $\theta_{DM} \gets \theta_{DM} + (r - r^*) \nabla \log p_{\theta_{DM}}(t | \mathbf{c}^{DM})$
                
                \Comment{update decision maker}
            \EndFor
        \EndFor
        
        \State $D \gets \texttt{KeepBestAndGiveUp}(D_c, H)$
        \State \textbf{train: } $C$ on $D$ using $L(\theta_C)$ \Comment{update phase}
    \EndFor

\EndProcedure
}
\end{algorithmic}
\label{algo:lifelonglearn}
\end{algorithm}

\begin{algorithm}[t!]
\caption{Interacting with the teacher}
\label{algo:capbyask}
\begin{algorithmic}[1]
{\footnotesize
\Procedure{SeekTeacher(I, greedy=False)}{}
    \State $\mathbf{w}^0$, $\mathbf{c}^0 \gets C(\cdot |I)$ \Comment{compute caption and context}
    \State $r^0 \gets \texttt{TeacherScore}(\mathbf{w}^0)$

    \For{$n=1$ to $N$}
        \State $t^n \gets DM(\cdot |\mathbf{c}^{DM, n-1}, \text{greedy})$ \Comment{DM samples step}
        \State $\mathbf{q} \gets Q(\cdot | I, c_{t^n}^{q, n-1})$ \Comment{generate question}
        \State $a \gets V(\cdot | I, \mathbf{q})$ \Comment{teacher provides answer}
        \State $\mathbf{w}^n_{ro}, \mathbf{c}^n \gets [\mathbf{w}^{n-1}_{0:t^n-1}, a, C(\cdot |I, h_{t^n-1}, a)]$
        \Comment{roll new cap.}
        \State $\mathbf{w}^n_{re} \gets [\mathbf{w}^{n-1}_{0:t^n-1}, a, \mathbf{w}^{n-1}_{t^n+1:}]$
        \State $r^n_{ro} \gets \texttt{TeacherScore}(\mathbf{w}^n_{ro})$
        \Comment{teacher scores caption}
        \State $r^n_{re} \gets \texttt{TeacherScore}(\mathbf{w}^n_{re})$
        \State $\mathbf{w}^n, r^n \gets \max \{r^{n-1}, r^n_{ro}, r^n_{re}\}$
        
    \EndFor
    
    \State \textbf{return} $r^N, \mathbf{w}^N$
    
\EndProcedure
}
\end{algorithmic}
\end{algorithm}

\subsubsection{Learning When to Ask Questions \label{sec:traindecisionmaker}}
As the agent queries the teacher, it trains the decision maker online to ask better questions. The teacher provides a scalar, non-differentiable reward. Hence the decision maker is updated using REINFORCE~\cite{sutton2000policy}. We baseline the reward with the greedy decision reward $(r^{*})^0$ (that is, what the improved-caption would have been had $DM$ sampled greedily), following the self-critical policy gradient~\cite{rennie2016self}. See line 11 in Algorithm \ref{algo:lifelonglearn}. In the general case where $N$ questions are asked, the gradient for the parameters of the decision maker $\theta_{DM}$ is:
{
\small
\begin{equation}
\label{eq:grad}
\sum_{n=1}^N [r^n - (r^{*})^n] \nabla \log p_{\theta_{DM}}(t^n | \mathbf{c}^{n-1})
\end{equation}
}
In this work we did not update the question generator in lifetime learning because jointly training the decision maker and question generator is a hierarchical RL problem. Reward accreditation is challenging because the agent needs to learn to differentiate DM choosing a bad time step from DM choosing a good time step but question generator generating a bad question.

\subsection{Captioner update phase\label{sec:traincaptioner}}
After the collection phase, the agent trains the captioning module on the collected captions. We assume the agent has full access to past data $D$ and is retrained from scratch. We retrain from scratch to avoid the added complexity of applying learning-without-forgetting techniques since our model has many moving parts already. Future works can look at how to efficiently learn on the new data. $D$ contains warmup GT captions, collected captions, and GT captions from ``giving up". The captioner is retrained using a joint loss over the captions stored in $D$,
{
\small
\begin{equation}
    \label{eq:loss}
L(\theta_C) = - \sum_{\mathbf{w}\in D} r_{\mathbf{w}} \log p_{\theta_C}(\mathbf{w} | I) - \lambda \sum_{\mathbf{w^*}\in D} \log p_{\theta_C}(\mathbf{w^*} | I)
\end{equation}
}
where $\mathbf{w}$ are collected captions, $\mathbf{w^*}$ are GT captions, $r_{\mathbf{w}}$ is the score given by the teacher for $\mathbf{w}$, and $\lambda$ is a tuned hyperparameter. In practice, we set $\lambda$ to the $90^{\text{th}}$ percentile reward of the collected captions, assuming that ground truth captions are generally better than collected captions.

\subsection{Implementation Details\label{subsec:modules}}
\paragraph{Captioning module.} $C( \mathbf{w} | I)$ is implemented as an attention CNN-RNN model~\cite{xu2015show}. We additionally predict a part-of-speech (POS) tag at each time step to inform the question generator what type of question should be asked and the decision maker whether to ask. Captioner is trained using MLE with teacher forcing and scheduled sampling. 

\paragraph{Question generation module.} $Q(\mathbf{q} | I, c^q_t)$ is also implemented as a CNN-RNN and conditions on the context at time $t$. Specifically, $c^q_t$ consists of: POS distribution which determines the ``question type", the attention weights predicted by the captioner which guide the question generator to look, an encoding of the caption which provides global context and prevents asking for redundant concepts, and the position encoding for $t$. We found it helpful to allow the question generator to re-attend rather than fully rely on the captioner's attention. We train the question generator on a novel dataset, using MLE with teacher forcing and scheduled sampling similar to the captioner (details in Appendix).

\paragraph{Decision module.}
The decision maker $DM(t | \mathbf{c})$ is implemented as a multilayer perceptron (MLP) with Softmax output.  Context $c^{DM}$ consists of the POS distribution, an encoding of the caption, and uncertainty metrics computed from top-k words predicted by the captioner:

\begin{itemize}[]
\item        Cosine similarity between the embedding of the top-1 word and all other $k-1$ words.\\[-4mm]
    \item Cosine similarity between each top-k word and the embedding of the entire sentence (implemented as the sum of word embeddings).\\[-4mm]
    \item Minimum distance of each top-k word to another word.\\[-3mm]
\end{itemize}
Entropy is a natural way to measure the uncertainty of the captioner. However, the model can predict synonyms which increase entropy but do not suggest that the model is uncertain. Therefore, for each time step we take the word embeddings of the top-k words and compute their relative distances as a secondary measure of uncertainty. We use $k=6$. In ablation studies, we show that these statistics alone can capture the uncertainty of the cap. Training a neural network on these stats further improves performance.

\paragraph{Teacher module.} We imagine our agent in a human-in-the-loop setting where a teacher answers natural language questions, chooses the best caption out of a few alternatives, scores it, and writes GT captions if necessary. The teacher consists of two parts: a VQA bot $V(a | I, \mathbf{q})$ implemented following~\cite{teney2017tips} and a caption scorer composed of a linear combination of BLEU \cite{papineni2002bleu}, ROUGE \cite{lin2004rouge}, METEOR \cite{banerjee2005meteor}, and CIDEr \cite{vedantam2015cider}. We call the reward from the caption scorer the \texttt{Mix} score, and denote it by $r$. We discuss challenges to using a synthetic teacher in Sections \ref{sec:learnbyaskq} and \ref{sec:understandthemodel}.
\section{Experiments}
\begin{table*}[t]
\begin{center}
\begin{tabular}{l c c c c c c c c c}
Method & $H \%$ &GT \% &Supervision \% &Mix 
&CIDEr
&METEOR
&ROUGE
&BLEU4
&BLEU2 \\
\hline\hline
$\text{Equal GT}$& - & 45.2 \% & 45.2 \% & 98.9 & 91.5 & 24.7 & 52.3 & 28.0 & 53.4  \\
$\text{All GT}$ & - & 100 \% & 100 \% & 101.7 & 96.4 & 25.1 & 52.9 & 28.8 & 54.9 \\
\hdashline
Inquisitive Student & 70\% & 45.2 \% & 73.5 \% & \textbf{103.9} & \textbf{98.0} & \textbf{25.4} & \textbf{53.8} & \textbf{30.5} &  \textbf{57.1} \\
Mute Student & 70\% & 45.2 \% & 72.6 \% & 102.2 & 95.9 & 25.2 & 53.4 & 29.3 & 55.9\\
\hline
\end{tabular}\\
\end{center}
\caption{Evaluation on \emph{test}. Our model was trained with 10\% warmup and 3 unlabelled chunks. Methods see all images at least once for fairness. \textbf{Note:} (Best of 3 runs) 100\% GT corresponds to 46\% of the MSCOCO training captions because only 2 (out of 5) captions are used for each image in the lifetime chunks.}\label{table:mainresult}
\end{table*}

\begin{figure*}[t]
\centering
\includegraphics[width=1.0\linewidth,height=4.5cm]{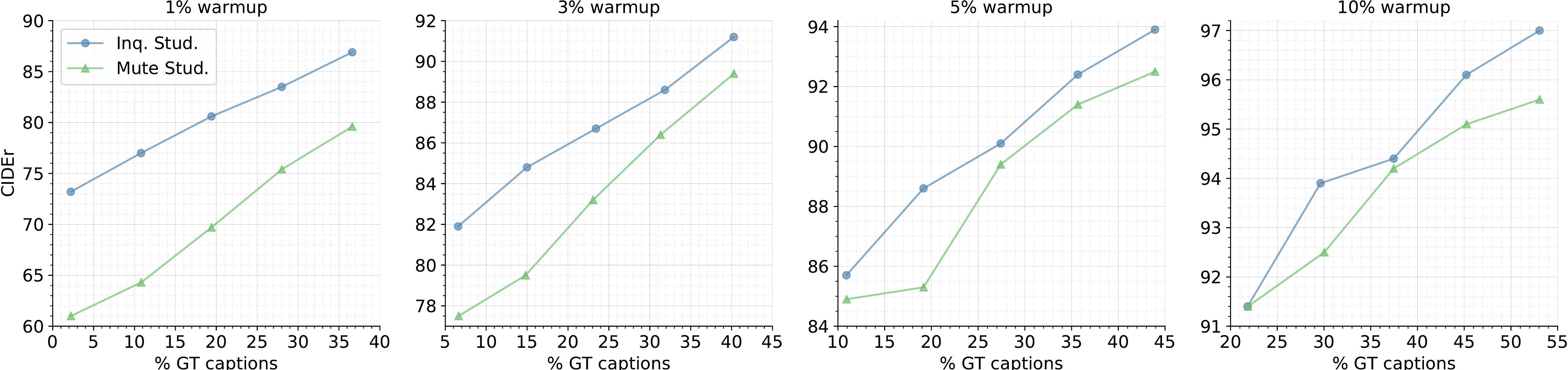}
\caption{Caption quality on test. Both models are decoded greedily.For each plot, GT \% is varied by changing the percentage of captions $H\%$ collected by the agent. \% GT captions is reported relative to \emph{All GT}.
}\label{fig:warmupplot}
\end{figure*}

We evaluate our approach on the challenging MSCOCO dataset~\cite{lin2014microsoft}, and compare it to intelligent baselines. We perform detailed ablation studies that verify our choices and give insight into how our model behaves. 

We follow the standard Karpathy split~\cite{karpathy2015deep} that contains 117,843 training, 5K validation and 5K test images. We randomly split the training set into warmup and lifetime learning chunks. In our experiments, we vary the size of the warmup, and the number of lifetime chunks, to analyze the model behavior under different regimes.  There are 5 GT captions for each image in the warmup set. At the end of lifetime learning, there are $m=2$ collected or GT captions for each image in the lifetime set. 

Image features are extracted with ResNet-101 trained on ImageNet~\cite{deng2009imagenet} \cite{he2016deep}. Vocabulary sizes for the captioner, question generator and VQA are 11253, 9755 and 3003, respectively. We use the Stanford NLP parser to get GT POS labels~\cite{manning2014stanford}. The decision maker only considers a subset of tags (listed in Appendix) for asking questions.

\subsection{Training Details\label{subsec:datasetdetails}}
The synthetic teacher (VQA bot) was trained on the VQA2.0 dataset \cite{antol2015vqa}, following a simplified implementation of \cite{teney2017tips} using a multi-answer binary cross entropy loss function. The VQA model achieves 64.2\% on the VQA2.0 val split without ensembling. We train the question generator by combining data from MSCOCO and VQA2.0. (Implementation details in App.) A natural concern is that training the question generator on images the captioner sees during lifetime learning will cause the que. gen. to ``lookup" GT questions. We find this to not be the case (see Figure~\ref{fig:selectqs}). In general, the questions generated for an image are diverse, generic and rarely match GT questions (see Appendix for more examples). The entire training process takes 2.5 longer than supervised learning baselines, mostly because we retrain the captioner from scratch. This slowdown can be overcome in future works by using learning-without-forgetting techniques.

\subsection{Cost of Human Supervision\label{sec:evaluation}}
We first perform a human study to understand human cost associated with every interaction type with the agent. We choose to measure ``human effort'' as the time taken for a task. In our experiment, a human teacher has three possible tasks: produce a full caption, answer a question, and score a caption. Table~\ref{table:interactionablation} shows that on average it takes 5.2 and 4.6 times longer to caption than score a caption or answer a question. To compute the cost of human supervision, we normalize the cost of each task to caption scoring. Hence the agent incurs one point of supervision for each caption scored, 1.13 for each question answered, and 5.2 for each caption written. In practice, we assume no cost when the VQA module answers a question. A human teacher would charge the agent for answers but would also give better answers. In the experiments to follow, we use \emph{Human Supervision} as a metric for cost incurred by querying a human.

\subsection{Learning by Asking Questions\label{sec:learnbyaskq}} 

In Table~\ref{table:mainresult} we evaluate our lifetime learner, \textit{aka} ``inquisitive student" (IS), against training only on GT data on the test split. All results are reported using greedy decoding. Our model was trained with a 10\% warmup chunk, 3 unlabelled chunks and 70\% collect percentage. For each setting we report the best model out of three with different random seeds. We report two GT baselines: \emph{Equal GT} -- the same number of GT captions as our model but fewer total captions, and \emph{All GT} -- the same number of captions as our model but only GT captions. 

\begin{figure*}[t]

\centering
\includegraphics[width=0.99\linewidth,height=6.2cm]{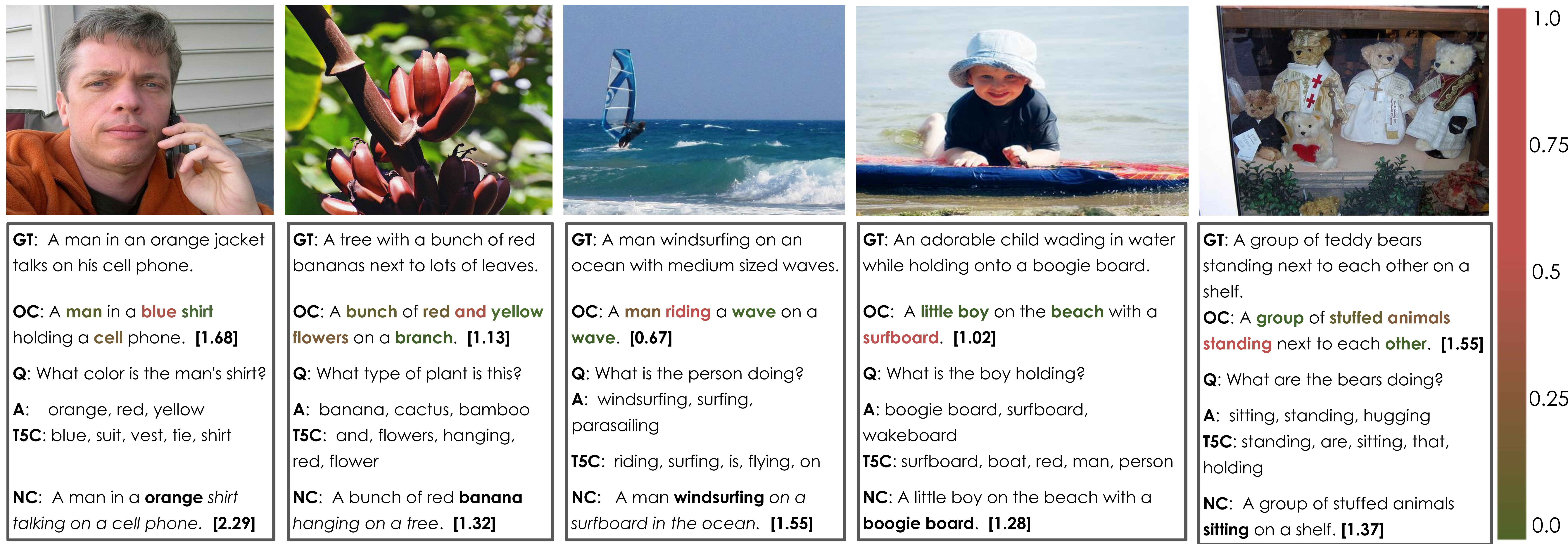}
\caption{T5C: top-5 words predicted by captioner at the word when question is asked. Rewards are in square brackets. Colors in OC indicate  probability the decision maker will ask about a word (scale is on right). Left $4$ are positive examples, right is failed (pointing to weaknesses of auto-eval metric). NC is the ``rollout" caption. Even when one word (answer) is replaced, multiple words can be updated because the captioner samples the rest of the sentence conditioned on the answer. 
}
\label{fig:examples}
\end{figure*}

\begin{figure*}

\begin{minipage}{0.33\linewidth}
\centering
\includegraphics[height=4cm]{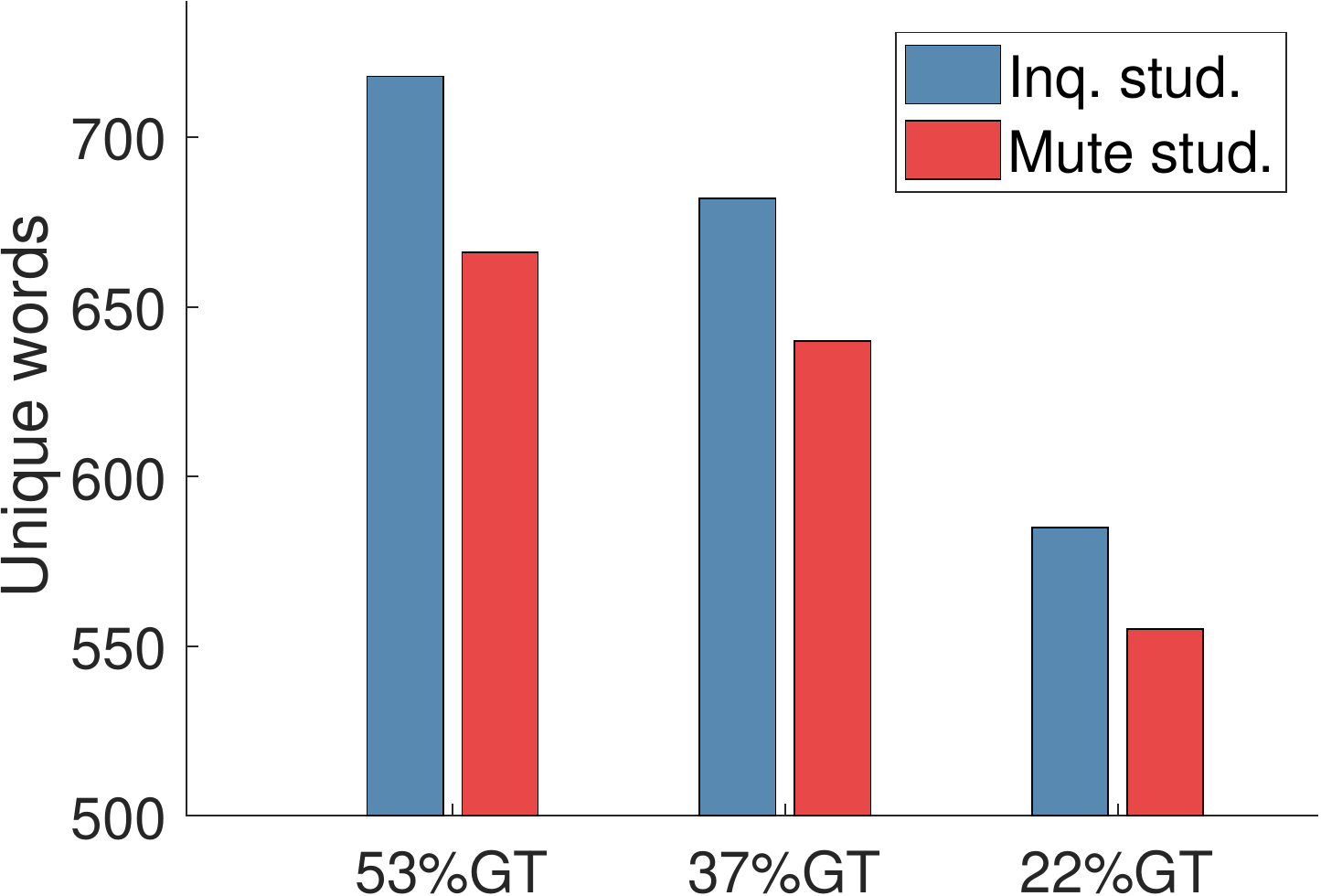}
\caption{Num. of unique words used by captioner evaluated on val at the end of lifetime learning. Models trained with 10\% warmup and 3 chunks.}
\label{fig:vocabplot}
\end{minipage}
\hspace{3.5mm}
\begin{minipage}{0.3\linewidth}
\centering
\includegraphics[height=4cm]{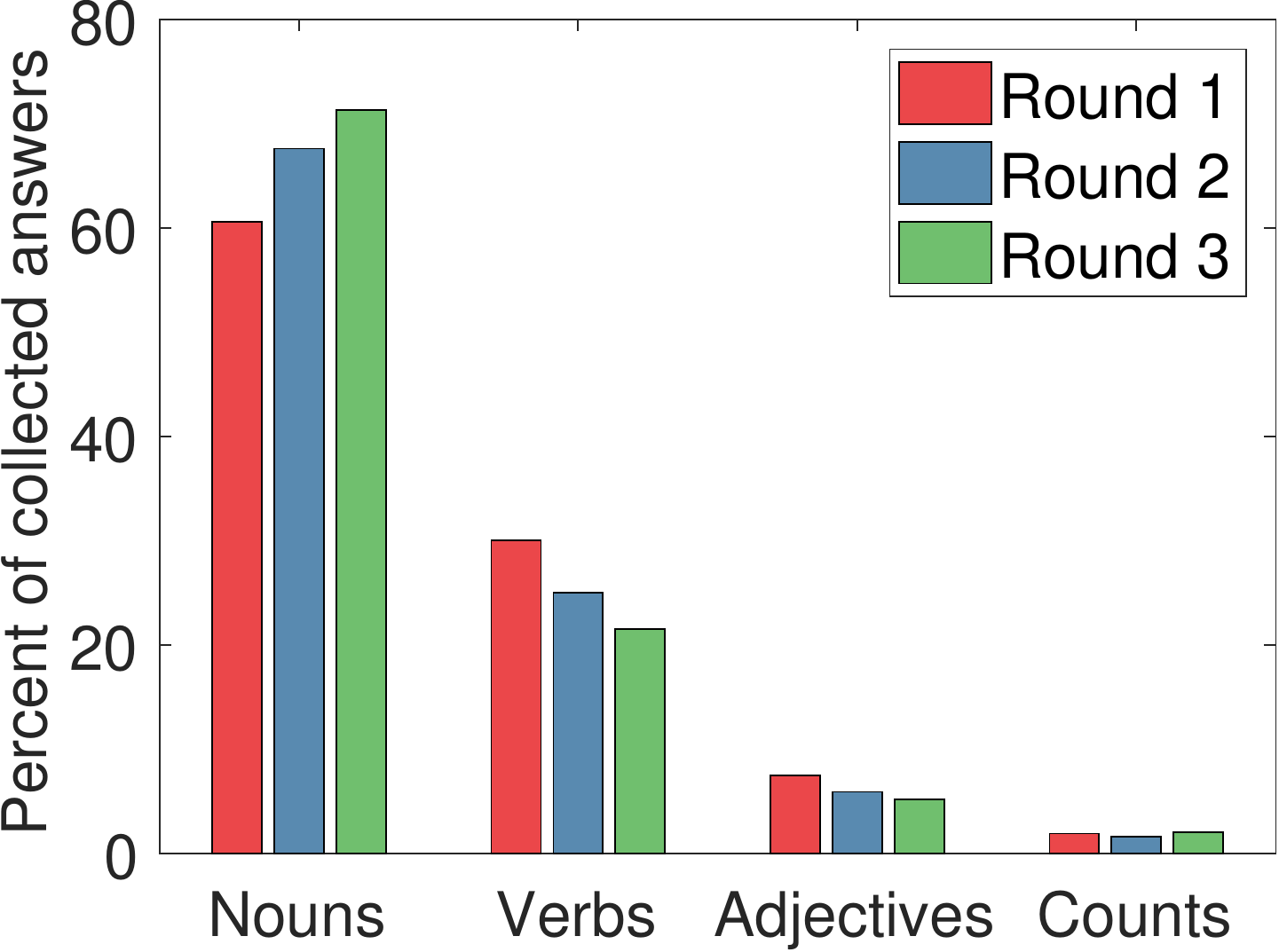}
\caption{Distribution of teacher answer types over rounds. The model was trained using 10\% warmup, $H=70\%$ and 3 chunks. }
\label{fig:answerdistrib}
\end{minipage}
\hspace{3.5mm}
\begin{minipage}{0.3\linewidth}
\centering
\includegraphics[height=4cm]{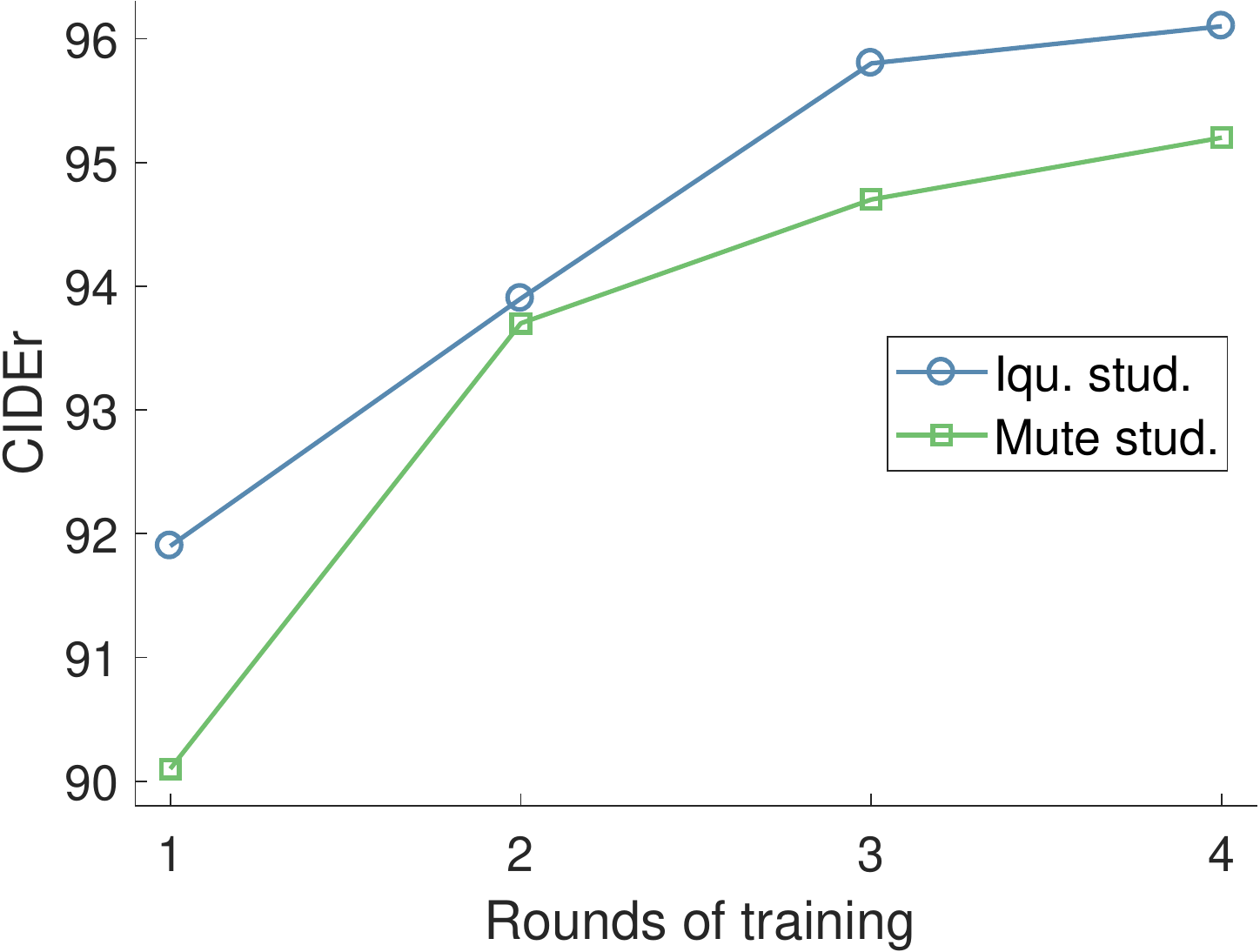}
\caption{Performance on val vs the number of total chunks (plus the warmup). Models were trained using 10\% warmup and $H=70\%$. }
\label{fig:performancevsround}
\end{minipage}
\end{figure*}

In order to evaluate the benefits of asking questions, we introduce Mute Student (MS), a lifetime learner that interacts with the teacher by only receiving feedback on whether captions are good (does not ask questions). MS is trained in exactly the same lifetime setting as IS, but samples multiple captions from the captioner's current distribution rather than ask questions to construct new captions to be rated by the teacher. The best captions are still collected and used to train for the next round. 
All models have the same hyperparameters and captioning architecture and are trained on all images to ensure fairness. GT and Supervision \% are reported relative to \emph{All GT}. 

Compared to \emph{Equal GT}, our lifetime model achieves 5 \texttt{Mix} and 6.5 CIDEr higher which shows that for an agent with a fixed budget of GT captions, additionally learning from collected captions can significantly improve performance. Compared to \emph{All GT}, our model achieves 2.2 \texttt{Mix} or 1.6 CIDEr higher score while using only 45.2\% of GT captions and 73.5\% of human supervision. This means that training on teacher-improved captions not only achieves greater efficiency but also leads to higher performance than training on GT captions. We find this to be a particularly strong and interesting result. 

IS also beats MS, which demonstrates that question-asking is beneficial. This is investigated further in Fig.~\ref{fig:warmupplot}. We vary the amount of GT captions by adjusting the percentage $H$ of collected captions. We call an agent that trusts its teacher-improved captions often (and rarely gives up) a ``confident" learner. Confident learners use less human supervision. An agent that begins lifetime learning earlier with only a small warmup set is an ``eager" learner. 

IS outperforms MS in almost all settings but the difference is greater if the agents are eager. Fig.~\ref{fig:warmupplot} shows that at 10\% warmup the gap is 1.4 CIDEr (97 vs 95.6) but as we reduce to 1\% warmup, the gap becomes 12.7 CIDEr (77 vs 64.3). This supports the intuition that asking questions benefits learners with less experience. In addition, a more eager learner ultimately reaches lower performance for the same amount of supervision. For about 30\% supervision IS achieves 93.9 CIDEr in the 10\% warmup setting and 83.5 CIDEr in the 1\% warmup setting. We hypothesize this is because the quality of sentence continuations, or rollouts after receiving the teacher's answer, worsens if the agent pretrains on less data. Furthermore, a very eager learner may make too many mistakes to fix by asking only one question. 

Selected examples are shown in Fig~\ref{fig:examples}. The first four examples are positive and show asking questions helps fix incorrect words and retrieve novel concepts.  In the fifth example, the reward is lower for the new caption even though it is good according to human judgment. Auto-eval metrics do not reward the agent for relevant, novel captions that don't match words in the reference captions. A human teacher with more flexible scoring could encourage the agent to learn more diverse captions and a larger vocabulary.
\subsection{Learning New Concepts\label{subsec:learnnewconcep}}
\begin{figure*}[t]
\centering
\includegraphics[width=1.0\linewidth,height=5cm]{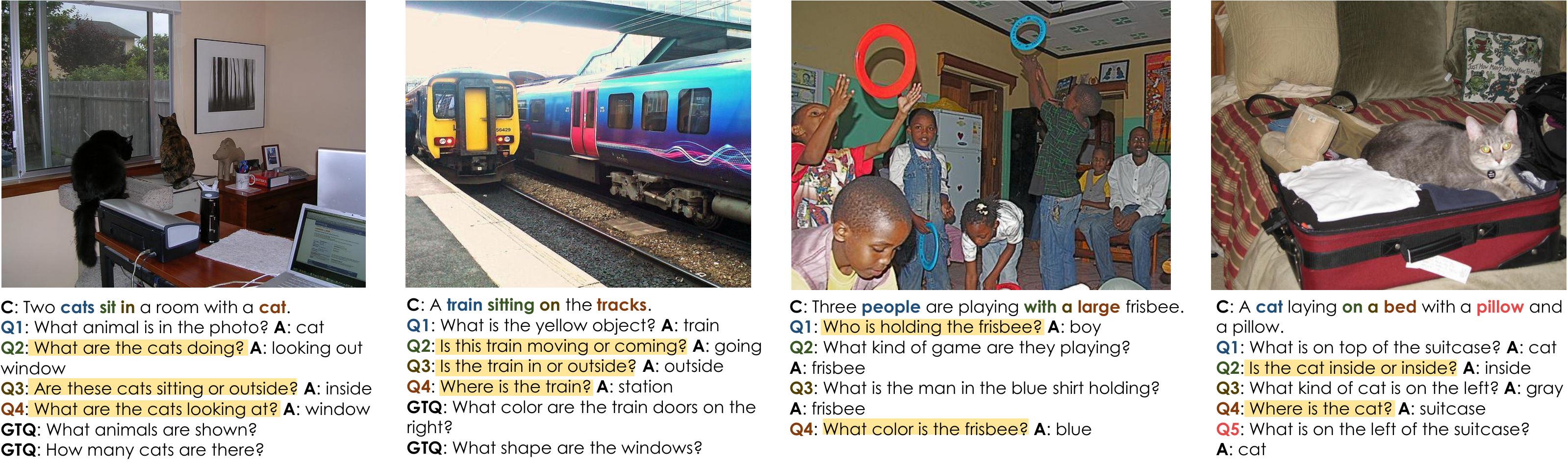}
\caption{Questions generated from different words in the generated caption (colors match words to questions). Highlighted questions retrieve answers that are novel to the caption. Left 2 images are seen by question gen. during training (GTQ are GT questions used for training), right 2 are not. Generated questions tend to be diverse and different from GT ones. }
\label{fig:selectqs}
\end{figure*}

\begin{table*}
\begin{minipage}[t]{0.291\linewidth}
\begin{center}
\begin{small}
\addtolength{\tabcolsep}{-2pt}
\begin{tabular}{l c c c}
Round & ATop3 & ATop5 & ATop10 \\
\hline
1 & 17.7  & 26.3  & 37.4  \\
2 & 24.1  & 34.2  & 46.9  \\
3 & 27.4  & 38.3  & 50.7  \\
\hline
\end{tabular}
\end{small}
\end{center}
\caption{Frequency (in \%) of teacher answers that occur in captioning module's predictions during lifetime training. Calculated from agent's collected captions in each round.}\label{table:ansintopk}
\end{minipage}
\hspace{1.5mm}
\begin{minipage}[t]{0.24\linewidth}
\begin{center}
\begin{small}
\addtolength{\tabcolsep}{-2pt}
\begin{tabular}{l c c c}
Model & Nouns & Verbs & Adj. \\
\hline
IS & 527 & 97 & 53 \\
MS & 491 & 86 & 48 \\
All GT & 680 & 127 & 47 \\
\hline
\end{tabular}
\end{small}
\end{center}
\caption{Number of unique words used by each model on val. Lifetime learners are trained with 10\% warmup, $H=60\%$, 3 chunks.}\label{table:vocabcap}
\end{minipage}
\hspace{1.5mm}
\begin{minipage}[t]{0.43\linewidth}
\begin{center}
\begin{small}
\addtolength{\tabcolsep}{-2pt}
\begin{tabular}{l c c c}
Task & Avg. time (s) & Std. (s) & Time ratio \\
\hline
Captioning & 34.4 & 21.8 & 1.0 \\
Scoring & 6.6 & 2.2 & 5.2 \\
Answering & 7.6 & 3.7 & 4.6 \\
\hline
\end{tabular}
\end{small}
\end{center}
\caption{Time taken by humans to perform tasks: captioning, scoring a caption, answering a question. Time ratio is relative to captioning. $N=27$ humans surveyed, $n_c=270$ captions written, $n_q=675$ questions answered, $n_s=675$ captions scored.}
\label{table:interactionablation}
\end{minipage}
\end{table*}

1\%, 3\% and 10\% warmup datasets contain only 30\%, 47\%, and 70\% of the captioning vocabulary respectively. The remaining words/concepts are explored in lifetime learning. Fig. \ref{fig:vocabplot} shows the number of unique words used by a captioner evaluated on the val split at the end of lifetime learning. We found a dependency between training epochs and vocabulary size and therefore took all models at the same epoch. We baseline against mute student. IS has a larger knowledge base than MS at all \% GT as it uses more unique noun, verb and total words than MS, showing IS is able to learn new vocabulary. 

In Table \ref{table:vocabcap} we compare the vocabulary of lifetime learners to \textit{All GT}. \textit{All GT} has a larger vocabulary than lifetime learners. This is intuitive because \textit{All GT} has more GT captions and therefore sees more varied data. IS only receives a single word answer given an image, whereas \textit{All GT} receives a complete caption label containing on average 10.5 words. For the same reason, in Fig. \ref{fig:vocabplot} the agents' vocabulary decreases as \% GT decreases.

Another way to measure the usefulness of teacher's answers is to compute how often it repeats a concept the captioner already knows. Table \ref{table:ansintopk} shows how frequently the answer from the teacher appears in the top-k words predicted by the captioner at the time step where the question is asked (ATopk). Note that this is approximate because the captioner may predict the answer at a different step. In the first round of lifetime training, 26.3\% of teacher answers appeared in the top-5 words predicted by the captioner. Hence, 73.7\% of the time, the agent is sees an unfamiliar or novel concepts. Over the lifetime, ATopk increases as the student's knowledge catches up to that of the teacher.  

\begin{figure*}[t!]
\begin{minipage}{0.35\linewidth}
\begin{center}
\addtolength{\tabcolsep}{-1.2pt}
\begin{tabular}{l c c c}
Method & Mix & C & B4 \\
\hline
No questions & 86.4 & 74.1 & 22.1 \\
\hdashline
Random & 88.3 & 76.2 & 22.2 \\
Entropy & 88.9 & 76.5 & 22.4 \\
Unc. metrics & 89.6 & 77.5 & 22.5 \\
\hdashline
Unc. metrics learned & 90.8 & 79.3 & 23.2 \\
Full learned & 91.9 & 80.6 & 23.7 \\
\hline
\end{tabular}
\end{center}
\caption{Ablating the decision maker. \textit{Entropy} is picking the time step with highest top-k word entropy. \textit{Unc. metrics} includes entropy and words closeness (Sec.~\ref{subsec:modules}). \textit{Unc. metrics learned} adds a MLP to predict the best time step for asking. \textit{Full learned} additionally includes POS and an encoding of the caption as input.}
\label{table:dmablation}
\end{minipage}
\hspace{4.5mm}
\begin{minipage}{0.26\linewidth}
\centering
\includegraphics[height=3.3cm]{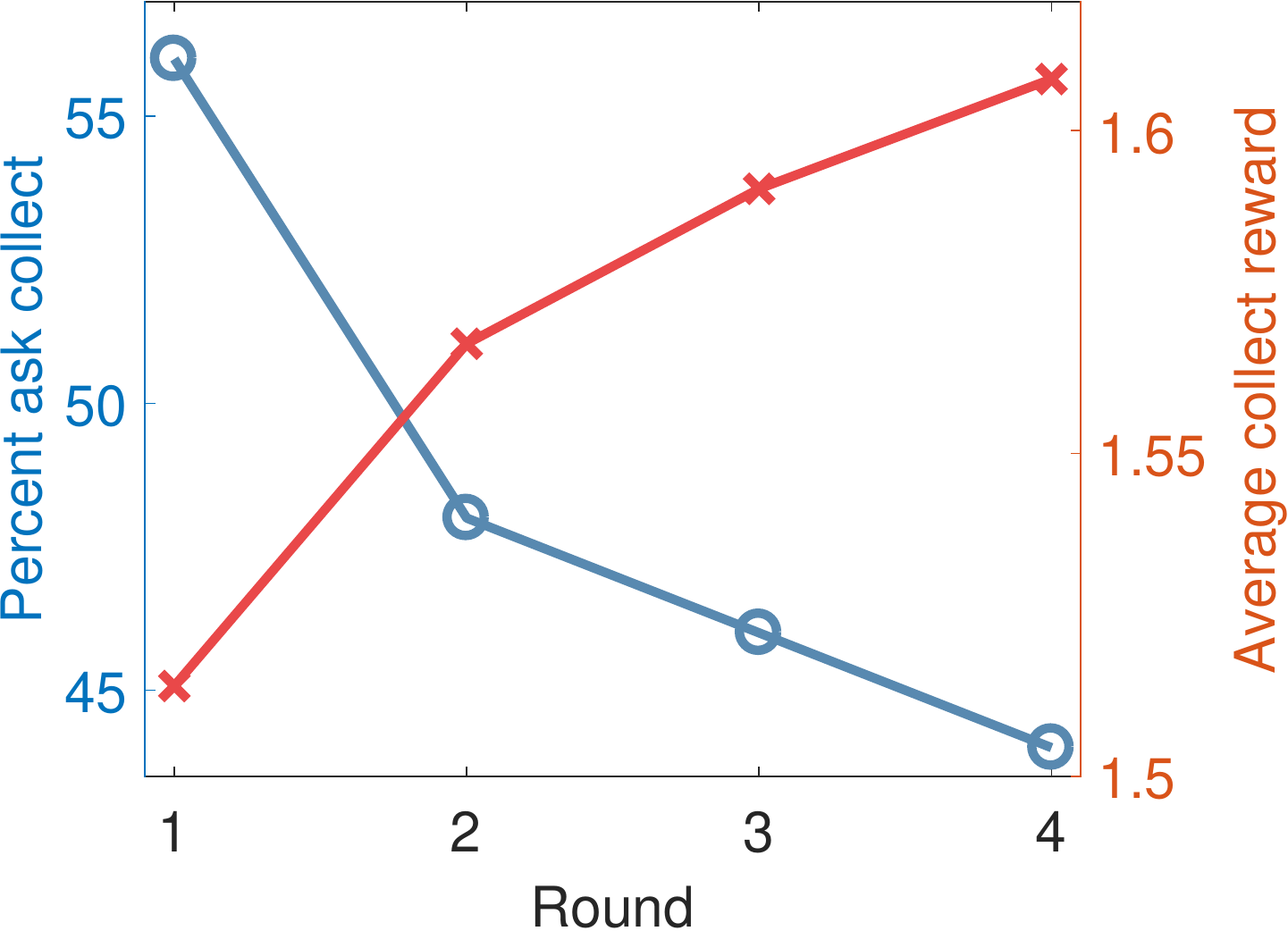}
\caption{Changes to collected captions over rounds. Model trained with 10\% warmup, $H=70\%$, 3 chunks.}\label{fig:roundablations}
\end{minipage}
\hspace{4.5mm}
\begin{minipage}{0.33\linewidth}
\centering
\includegraphics[height=2.4cm]{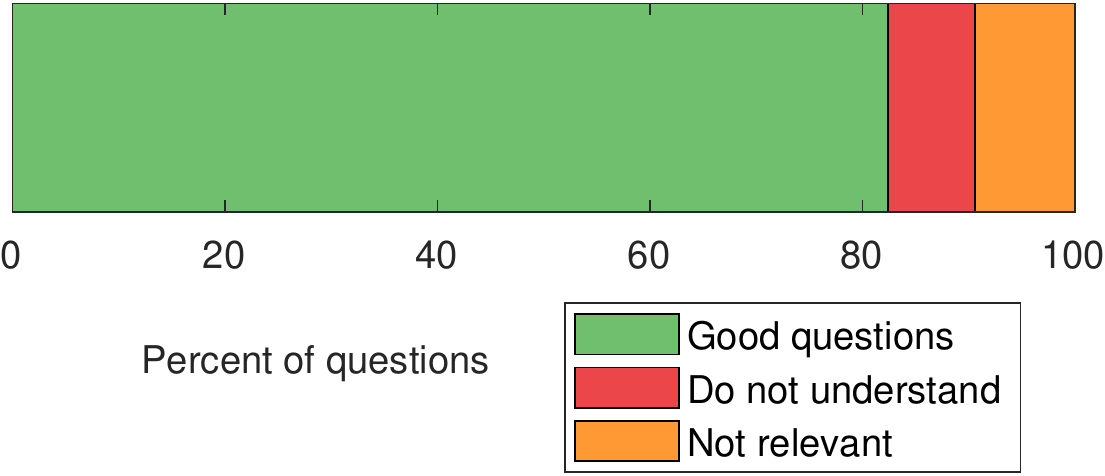}\\[1.6mm]
\caption{AMT study to judge the quality of the generated questions. Given an image and a question, annotators were asked to answer the question if it is good, or flag it as ``not understandable" or ``not relevant". Generally the questions were good.}
\label{fig:questionquality}
\end{minipage}
\end{figure*}

\subsection{Analyzing the Modules\label{subsec:analymodules}}
\paragraph{Question Generator.} We conducted a human study (Fig.~\ref{fig:questionquality}) using Amazon Mechanical Turk (AMT) to evaluate the quality of generated questions. Annotators rated 500 images-question pairs by answering questions if they were good or flagging questions as ``not understandable" or ``irrelevant to the image". The questions were randomly selected questions that the question generator asked while trying to caption. The images were not seen by the question generator during its training. 82.4\% of questions were rated ``good" and answered. This is a promising result and suggests that learning by asking can be adapted to use human teachers instead of a QA bot. 

Fig.~\ref{fig:selectqs} shows generated questions at different time steps in a caption. In general, generated questions tend to be diverse, and generic. It's important for questions to be generic so that the teacher can answer with a wide range of possible concepts and possibly new concepts. We also rarely observe the generated questions to be the same as the GT questions. More examples in Appendix.

\paragraph{Decision Maker.} To test the decision maker, we look directly at the scores of the refined captions it produces, rather than those of the final captions after retraining the captioner. This lets us to precisely observe the ablated performance of the DM. Table ~\ref{table:dmablation} evaluates different decision maker strategies. We first train captioning and question generation modules. The baseline is the performance of the captioner without asking questions. The other settings use various decision maker models to ask a question to improve captions. Learned models are trained using RL on a single chunk of unlabelled data. Scores are shown for the \textit{val split}. 

The full model gives 6.5 CIDEr improvement over no question asking. Picking the time step with maximum entropy is not a very good strategy. It is only 0.3 CIDEr better than picking a random step. This is because the model can predict synonyms which increase the entropy but do not indicate the model is uncertain. Adding closeness metrics yields 1.0 CIDEr improvement over maximum entropy, showing that taking into account the closeness of words in embedding space gives a better measure of uncertainty. In all cases, learning improves performance, with the best learned model achieving 3.1 CIDEr higher than the best non-learned model. We use the full model as our decision maker for all experiments.

\subsection{Understanding the Model\label{sec:understandthemodel}}

\paragraph{Number of chunks.} Fig. \ref{fig:performancevsround} shows that as the number of chunks increases, performance increases (for similar human supervision). This is intuitive because more chunks means the agent sees fewer images before adapting the captioner. The number of chunks cannot be too large because we retrain the captioner from scratch after every chunk.

\paragraph{Catching up to the teacher.} Fig.~\ref{fig:roundablations} shows the percent of collected captions that improved by asking questions (left axis) and average reward of collected captions (right axis) versus num. consumed chunks. Over time, the agent is able to improve fewer and fewer captions by querying the teacher. Furthermore, the largest increase in collected reward occurs in the first round. These observations suggest that the teacher's knowledge is exhausted over time. This is a limitation of using a static, synthetic, and noisy QA-bot (which only achieves 64\% accuracy). Learning may benefit from human teachers over more rounds, because they are more accurate and have a much wider pool of  knowledge.

\paragraph{Types of answers.} In Fig.~\ref{fig:answerdistrib} we see the distribution of answer types from the teacher. Over time, the student asks for more nouns, and less verbs and adjectives. We hypothesize this is because the agent is learning verbs and adjectives early on before moving onto nouns.
\section{Conclusion}
In this paper, we addressed the problem of active learning for the task of image captioning. In particular, we allow the agent to ask for a particular concept related to the image that it is uncertain about, and not require the full caption from the teacher. Our model is composed of three modules, \ie captioning, decision making and question posing, which interact with each other in a lifetime learning setting.
Done this way, the learning and teaching efficiency is shown to be improved on the challenging MS-COCO dataset. 

Our work is the first step towards a more natural learning setting in which data arrives continuously, and robots learn from humans through natural language questions and feedback. There are many challenges ahead in making the lifetime model learning more efficient, and incorporating real humans in the loop. 

\paragraph{Acknowledgements} Supported by the DARPA Explainable AI (XAI) program. We thank NVIDIA for their donation of GPUs. We thank Relu Patrascu for infrastructure support,  David Acuna, Seung Wook Kim, Makarand Tapaswi, Yuan-Hong Liao, for fruitful discussion and Atef Chaudhury, Harris Chan, Silviu Pitis for their helpful feedback in editing the paper.

{\small
\bibliographystyle{ieee}
\bibliography{egbib}
}

\clearpage

\section{Supplementary Material}

This supplementary contains details of the modules in our model, the training procedure, as well as additional qualitative examples. In Section~\ref{sec:AppImpDet} we discuss implementation details of the captioner, question generator, decision maker and VQA teacher. Furthermore, we describe the inquisitive student and mute student in the lifetime setting. In Section~\ref{sec:AppAsk} we discuss the challenges with asking $N>1$ questions. In Section~\ref{sec:AppHum} we provide more detail on how human supervision is calculated for our experiments. In Section~\ref{sec:AppQAbl} we show an ablation study on the question generator. Our study highlights the importance of each feature in the context used by the question generator. In Section~\ref{sec:AppExam} we show more qualitative examples and describe the failure modes of our model.

\subsection{Implementation Details \label{sec:AppImpDet}}

\subsubsection{Lifetime Learning}

In lifetime learning, data arrives in chunks. In the collection phase, the agent attempts to improve generated captions by querying the teacher. In the update phase, the captioning module is updated on collected captions.

In our experiments, we vary the collection percentage $H\%$ and the size of the warmup chunk. Note: the size of the warmup chunk is reported relative to the entire training split whereas \% GT (reported in tables and figures) is relative to the total number of captions the baseline \textit{All GT} is trained on. For example 10\% warmup refers to a dataset with 11.3K/113K images and 57K/567K captions. We explored the following settings.

\begin{itemize}
    \item $H\%$: 60, 70, 80, 90, 100\%
    \item warmup: 1, 3, 5, 10\%
\end{itemize}

In the update phase, we train the captioner with ADAM \cite{kingma2014adam}, \texttt{lr=$2\mathrm{e}{-4}$}, \texttt{batchsize=$20$}, scheduled sampling, and learning rate (lr) decay. Scheduled sampling and learning rate decay are described in \ref{sec:suppcapmod}. We now outline details of inquisitive student (IS) and mute student (MS) in the collection phase.

\paragraph{Inquisitive student}
The inquisitive student samples captions and questions greedily. We found it helpful to put the captioner and question generator into \texttt{eval} mode so that dropout is still applied. This introduces a small amount of stochasticity and the agent generates more varied captions. QE makes 8 passes over each image in a chunk. However, because the captioner and question generator are sampled greedily, later rounds only produce a few novel captions. We found that 4 passes worked almost as well; more passes produces diminishing returns. We train the decision maker online using policy gradient. We use ADAM, \texttt{lr=$2\mathrm{e}{-5}$} and \texttt{batchsize=$20$}

\paragraph{Mute student}

Mute student has the same captioning architecture and hyperparameters as QE. There are some differences in the collection phase. Instead of asking questions, ME samples from the captioning module to explore new captions. Specifically, ME samples captions with temperature 1.0. ME makes 4 passes over each image in a chunk. This is to ensure that QE and ME use similar amount of human supervision.

\subsubsection{Captioning Module\label{sec:suppcapmod}}
The captioning module is implemented as an attention encoder-decoder. It predicts both the next word and the next POS given the previous word. In our implementation, the CNN encoder is fixed. However, we project image features using a fully connected (FC) layer before passing it to the decoder. The decoder is implemented as a single layer GRU with 512 units. We use \texttt{dropout=0.5} in all layers.

\paragraph{POS prediction}
The hidden state of the GRU is used to compute the next word and POS. More specifically, we first predict the POS distribution then condition the next word on either the predicted POS or ground truth POS. Scheduled sampling is used to control how often the predicted or GT POS is used. Words are embedded into 512 dimensional latent space and POS are embedded into 50 dimensional latent space. If the predicted POS is used to predict the next word, we embed the entire POS distribution and concatenate this embedding with the decoder hidden state. The resulting vector is passed into a FC layer to predict the next word. If the GT POS is used, we embed the one-hot vector and similarly predict the next word. The captioner is trained using a joint loss over next word and POS.

\begin{equation}
\label{eq:total loss}
L = L_\text{word} + \alpha L_\text{POS}
\end{equation}

We tune $\alpha$ and find $0.5$ to work the best. We limit the length of captions to 16 plus the end-of-sentence symbol.

\paragraph{Getting ground truth POS}
We use the Stanford NLP parser to get ground truth POS for both GT and collected captions. On rare occasions, the Stanford NLP parser returns errors when parsing generated captions. In these cases, the agent collects the GT caption for the image rather than the generated one.

\paragraph{Training}
Training is the same for both the warmup chunk and update phase in lifetime learning. Specifically, we train the captioner using MLE with teacher forcing, scheduled sampling (on both the words and POS) and learning rate decay. We start the learning rate at $2\mathrm{e}{-4}$ and decays it by 0.8 every 3 epochs. The probability of predicting the next word using the \textit{previous predicted word} (rather than the GT word) starts at 0 and increases by 0.05 every 5 epochs. The probability of predicting the next word using the \textit{previous predicted POS} (rather than the GT POS) starts at 0.2 and increases by 0.05 every 3 epochs.

\subsubsection{Question Generating Module}

The question generator generates a question given a context vector computed by the captioner. More specifically, the context consists of the full caption and the POS and attention weights (of the captioner) at a particular time step. The time step is determined by the decision maker.

\paragraph{Encoding the caption}
The caption is used by the question generator in two ways. First, a pretrained (and fixed) captioning module is used to encode the caption. The hidden state of the captioner GRU (\textit{aka} cap-hid) is used as a feature. Second, the question generator encodes the caption with its own single layer Bidirectional-GRU (BiGRU). The BiGRU has 256 units. Caption words are represented as 256 dimensional latent vectors. Finally, the time step computed by the decision maker is encoded as a 256 dimensional vector of 1's. It is fed alongside the caption word embeddings into the BiGRU.

\paragraph{Architecture}

The question generator's decoder is also a single layer GRU with 512 units. We embed the POS and use it along with cap-hid to initialize the decoder hidden state. We use the entire POS distribution rather than the max POS. The BiGRU encoded caption is passed into the decoder along with image features at every time step. We limit the length of questions to 14. Question words are embedded into 512 dimensional latent space.

\paragraph{Re-attention}

We allow the question generator to re-attend to the image. Specifically, the question generator first computes its own attention weights independent of the captioner's weights. The attended image features are then concatenated with features computed from the captioner's attention. Finally a FC layer is used to compute the final image features.

\paragraph{Training}

We train the question generator with MLE, teacher forcing, learning rate decay and scheduled sampling. We use a batch size of 20 and the same schedules as for training the captioner: lr decay 0.8 every 3 epochs, scheduled sampling increase 0.05 every 5 epochs.

\subsubsection{Dataset for Training Question Generator}

We combine the MSCOCO and VQA2.0 datasets to train the question generator. The two datasets share images. Therefore, we can form training samples by matching answers from QA pairs of VQA2.0 to words in the MSCOCO captions. A training sample is a (caption, answer, question) tuple. We pass the caption through a pretrained captioner to compute the context vector. The ``time step" is chosen to be the index of the word in the caption that matches the QA answer. The question generator is trained to predict the GT question given the context. Doing this gives us 135K samples for training and 6K for validation. We call this the ``answer-matched" set. We make a second ``pos-matched" dataset to increase the diversity of questions by taking the answer from QA and instead matching its POS to a random word in MSCOCO captions with the same POS. The pos-matched dataset contains 108K samples. When we train the question generator, we sample from both the answer-matched and pos-matched datasets (equally) in every minibatch. \\

To make the VQA vocabulary match the captioning vocabulary better, we convert numbers from digits to words (\eg 7 $\rightarrow$ ``seven"). For every image in the answer-matched dataset, we allowed at most 2 questions \textit{with the same answer}. This is to prevent the model from overfitting and asking questions about only a single concept in an image.

\subsubsection{Decision Making Module}

The decision maker predicts which word in the generated caption the agent should ask about. It does so given a context vector from the captioner. The context vector consists of: the full caption, POS, attended image features, and the top-k words. The attended image features are computed by weighting the output of the CNN encoder by the captioner's attention weights. The top-k words are the top-k words predicted by the captioner at every time step. They are used to compute closeness metrics which capture the captioner's uncertainty. We use $k=6$.

\paragraph{Encoding the caption}

The decision maker encodes the full caption the same way as the question generator. First we pass the caption through a pretrained captioner. The hidden state of the GRU is used as a feature. Second, we encode the caption with a BiGRU with 256 units. 

\paragraph{Masking out invalid POS}

We mask out invalid POS (corresponding to words the agent can never ask questions about). We do this by computing the maximum of the POS distribution at each time step and comparing it to a predefined list of valid POS. See table \ref{table:poslist} for the full list.

\paragraph{Closeness metrics}

The top-k words predicted by the captioner are used to compute word closeness metrics. First words are embedded using the embedding layer from the captioner. Then, we compute the following features.

\begin{itemize}[leftmargin=*]
\vspace{-2mm}
\item        Cosine similarity between the embedding of the top-1 word and all other $k-1$ words.\\[-6mm]
    \item Cosine similarity between each top-k word and the embedding of the entire sentence (implemented as the sum of word embeddings).\\[-6mm]
    \item Minimum distance of each top-k word to another word.\\[-6mm]
\end{itemize}

The result is a $B \times T \times X \times k$ tensor where $B$ is the batch size, $T$ the number of time steps, $X$ the number of channels/features. and $k=6$. We combine the features along the $k$ axis using a 1D CNN. The final feature vector is a $B \times T \times 512$ tensor.

\paragraph{Computing the probability of asking questions}

We embed the POS distribution with an embedding layer. We project the image features with a FC layer. Finally we pass the POS, image, caption and closeness features through a MLP to compute logits. We apply a \textit{Softmax across time} to find the probability of asking a question at each time step.

\begin{table}
\vspace{-3mm}
\begin{center}
\begin{small}
\begin{tabular}{l c}
POS & Description \\
\hline
NN & noun, singular \\
NNS & noun, plural\\
NNP & proper noun\\
NNPS & proper noun, plural\\
VB & verb\\
VBG & verb, gerund/present participle\\
VBD & verb, past tense\\
VBN & verb, past participle\\
VBP & verb, singular present\\
VBZ & verb, 3rd person singular present\\
JJ & adjective\\
JJS & adjective, superlative\\
JJR & adjective, comparative\\
RB & adverb\\
RBS & adverb, superlative\\
RBR & adverb, comparative\\
RP & particle\\
CD & cardinal number\\

\hline
\end{tabular}\\
\end{small}
\end{center}
\caption{\footnotesize{List of valid POS for the decision maker. All other POS are classified as ``other".}}\label{table:poslist}
\end{table}

\subsubsection{VQA}
We use a VQA model as a synthetic teacher-answerer. We remove $\texttt{yes/no}$ questions from the VQA dataset as they would not be useful for our captioning regime. This leaves 277K/444K questions in the training set and 133K/214K questions in the validation set.

\paragraph{Architecture}

We use a similar architecture to \cite{teney2017tips} but with \texttt{PReLU} activations instead of gated Tanh. We use \texttt{dropout=0.5} in all layers. Question words are encoded using an embedding layer of 300 dimensions. The word embedding is initialized using glove embeddings \cite{pennington2014glove} but we did not find significant difference versus training from scratch. We limit questions to 14 words. The full question is passed through a single layer GRU with 512 units to get a vector representation. We use the final hidden state (step 14) of the GRU (padding short sentences) without sequence trimming or dynamic unrolling. Captions are used as supporting evidence to increase the performance of the VQA model. Captions are also embedded using an embedding layer with 300 units and then encoded using a GRU. 

To fuse the question, image, and caption, we do an element-wise product between their vector representations. Specifically, we multiply question and image together, as well as question and caption together. The two feature vectors are concatenated and fed through a MLP to predict the logits.

\paragraph{Training}

We use \texttt{batchsize=$256$}, \texttt{lr=$1\mathrm{e}{-3}$} and ADAM to train the VQA model. Batch normalization is used to normalize image features. We use a multianswer loss. The loss for a single sample is shown below.

\begin{equation}
\label{eq:total loss}
L = - \sum _i ^M p_{i} \log (\hat p_{i}) - (1-p_{i}) \log (1-\hat p_{i})
\end{equation}

Here $i$ indexes over the answer vocabulary. $M$ is the size of the vocabulary. $p_i$ is the ground truth probability of answer $i$. $\hat p_i$ is the probability predicted by the model. Each question in the VQA2.0 dataset is answered by 10 humans. This loss takes the full empirical human answer distribution as the target rather than only the most common answer.

\subsection{Asking Multiple Questions\label{sec:AppAsk}}
In our reported experiments, the agent asked $N=1$ questions for each generated caption in the collection phase. We experimented with asking $N>1$ questions. However, this is challenging because the teacher's answer is directly inputted into the captioner to roll out the rest of the new sentence. If the answer is a rare or out of vocabulary word, the final words of the sentence may not follow a sensible trajectory. This problem is worsened when multiple questions are asked. One possible solution is to exploit hypernyms to keep the captioner on a sensible trajectory while inserting novel words. Another solution may be to learn an answer ``absorption" module to utilize the teacher's answer better. We leave these directions to future works.

\subsection{Calculating Human Supervision\label{sec:AppHum}}
In our reported experiments, we computed the cost of human supervision by considering the completion time of various tasks. More specifically, every GT caption a model has access to costs 5.2 units of supervision and every caption scored during lifetime learning costs 1 unit of supervision. We make some other assumptions when calculating human supervision.

First, we filtered out repeated captions and questions. Furthermore, we assume no cost when the VQA module answers a question. A human teacher would charge the agent for answers but would also give better answers. Finally, we only charge the agent once for picking the caption with the highest reward from the three alternatives: rollout, replace, and original and then scoring it. This assumption can be relaxed by training a caption discriminator in future works.

\subsection{Ablating the Question Generator\label{sec:AppQAbl}}

\begin{table}
\begin{center}
\begin{tabular}{|l|c|c|c|c|}
\hline
Method & a@1 & a@3 & a@5 & a@10 \\
\hline
Baseline & 37.8 & 50.2 & 55.3 & 62.7 \\
+CE & 45.9 & 60.2 & 65.5 & 72 \\
+PE & 49.2 & 63.9 & 69.3 & 75.4 \\
+PE +CE & 52 & 67.2 & 73 & 79.4 \\
\hline
\end{tabular}
\end{center}
\caption{Comparing question generation models using different context inputs. (+PE) with position encoding, (+CE) with RNN encoding of the caption.}
\label{table:qgenablation}
\end{table}

In table ~\ref{table:qgenablation} we show how including various features affects the accuracy of the question generator. We use accuracy as a proxy to question quality. Accuracy is measured by passing a generated question through the VQA module and comparing the teacher's answer with the ground truth answer. a@n means the GT answer appears in the top-n teacher answers. The baseline is a model trained only with POS and an attention maps as context. Results are reported on the validation split of the dataset used to train the question generator. Both position and caption encoding give a boost to the accuracy. Using both achieves 14.2\% accuracy over baseline. We use the full model as our question generator in experiments. 

\subsection{Qualitative Examples\label{sec:AppExam}}

\begin{figure*}[t]
\centering
\includegraphics[width=0.8\linewidth]{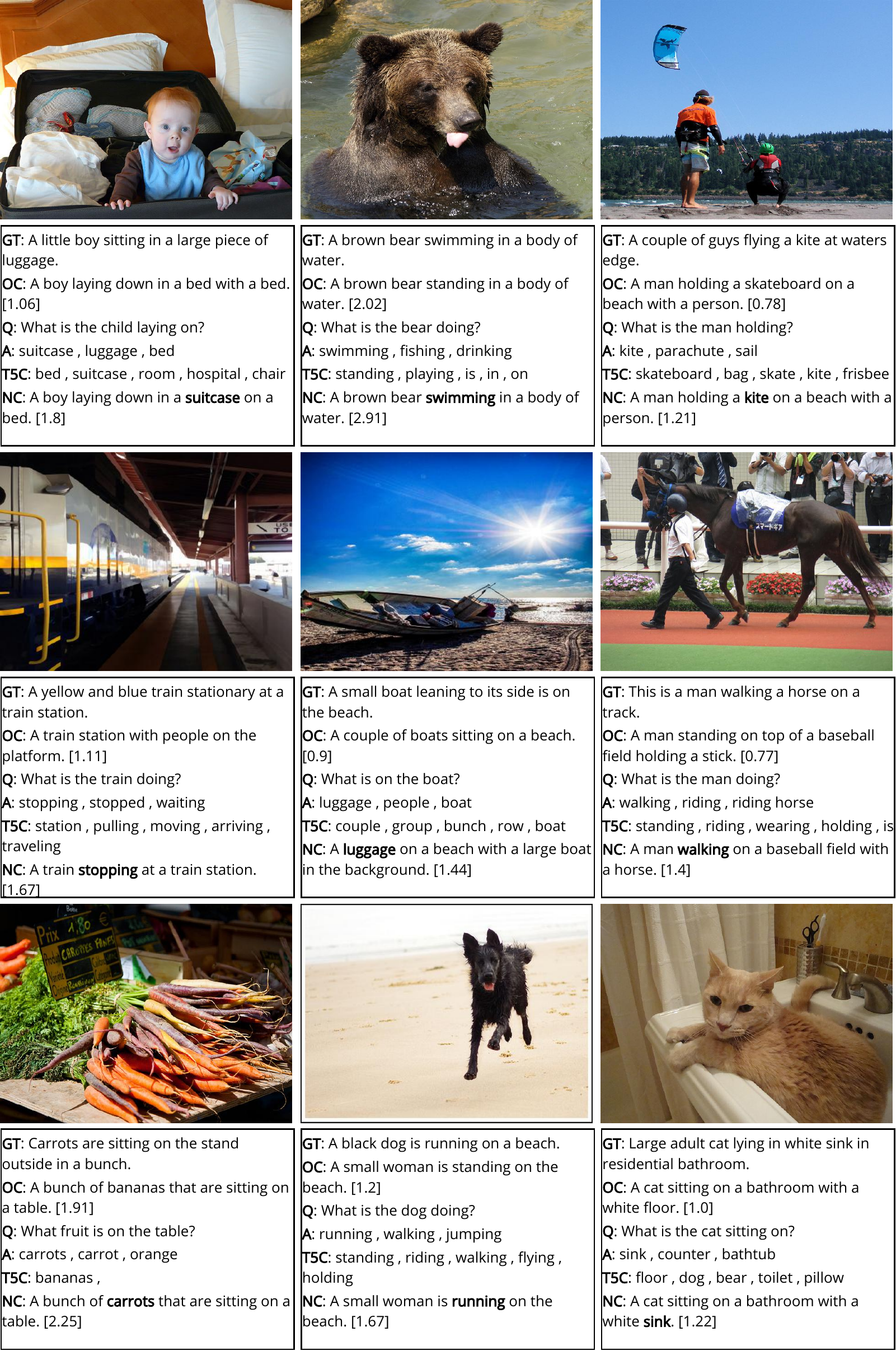}
\caption{\footnotesize Examples of the agent interacting with the teacher in the collection phase. T5C: top-5 words predicted by captioner at the word when question is asked. Rewards are in square brackets. First two rows show warmup dataset size of 10\%. The last row shows warmup dataset size of 1\%. For 1\% warmup, some generated captions (OC) have too many errors to fix with a single question.}
\label{fig:AppCExam}
\end{figure*}

\begin{figure*}[t]
\centering
\includegraphics[width=0.8\linewidth]{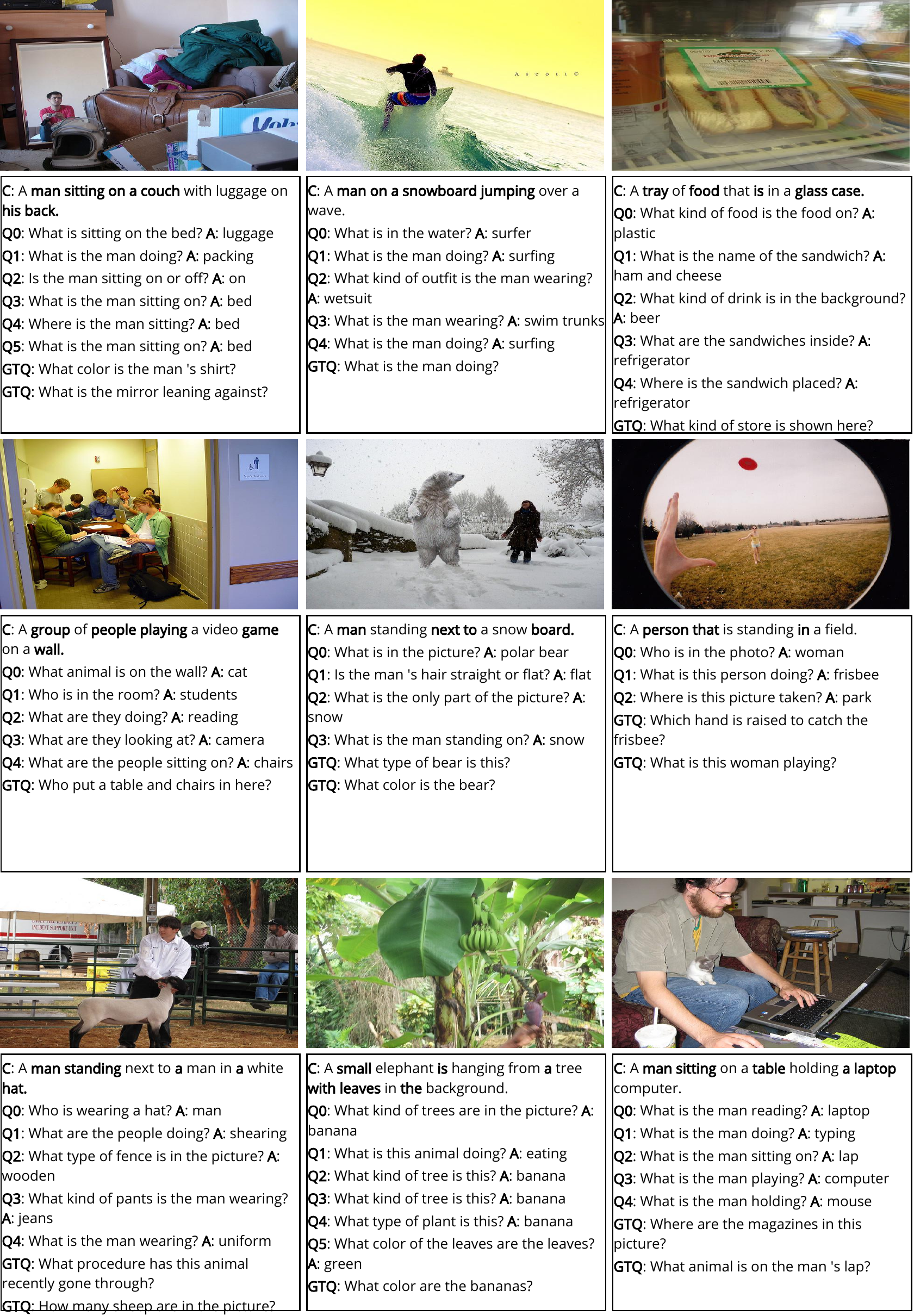}
\caption{\footnotesize Examples of generated questions at different time steps of a generated caption. The images were used to train the question generator. Not all generated questions are useful. This shows the importance of learning the decision maker to decide when and whether to ask questions. Up to six generated questions are shown. GTQ are ground truth questions used to train que. gen. Generated questions are almost always different from GT ones. Questions are asked for bolded words in the caption. The order of questions corresponds to the order of bolded words. \ie Q0 corresponds to the first bolded word, Q1 corresponds to the second bolded word, and so on. The caption is generated from a model trained on 10\% warmup data.}
\label{fig:AppQExam2}
\end{figure*}

\begin{figure*}[t]
\centering
\includegraphics[width=0.8\linewidth]{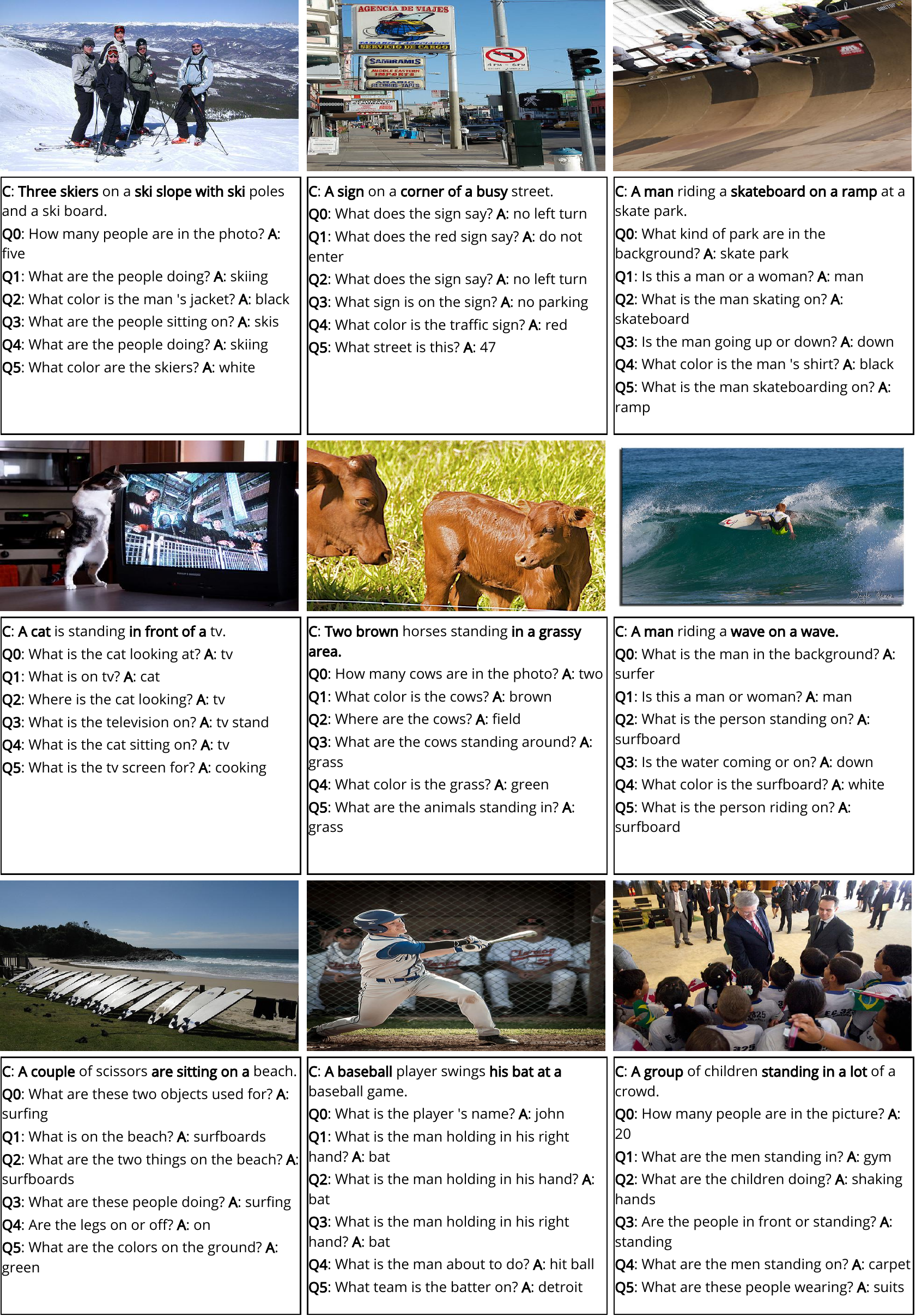}
\caption{\footnotesize Random examples of generated questions at different time steps of a generated caption. The images were (unseen) \textbf{not} used to train the question generator. Questions tend to be diverse and generic. Up to six generated questions are shown. Questions are asked for bolded words in the caption. The order of questions corresponds to the order of bolded words. \ie Q0 corresponds to the first bolded word, Q1 corresponds to the second bolded word, and so on. The caption is generated from a model trained on 10\% warmup data.}
\label{fig:AppQExam1}
\end{figure*}

More qualitative examples are shown in the following pages. Fig. \ref{fig:AppCExam} shows the agent interacting with the teacher in the collection phase of lifetime learning. Figs. \ref{fig:AppQExam2}  and \ref{fig:AppQExam1} show generated questions.

\subsubsection{Failure Modes}

\begin{figure*}[t]
\centering
\includegraphics[width=1.0\linewidth]{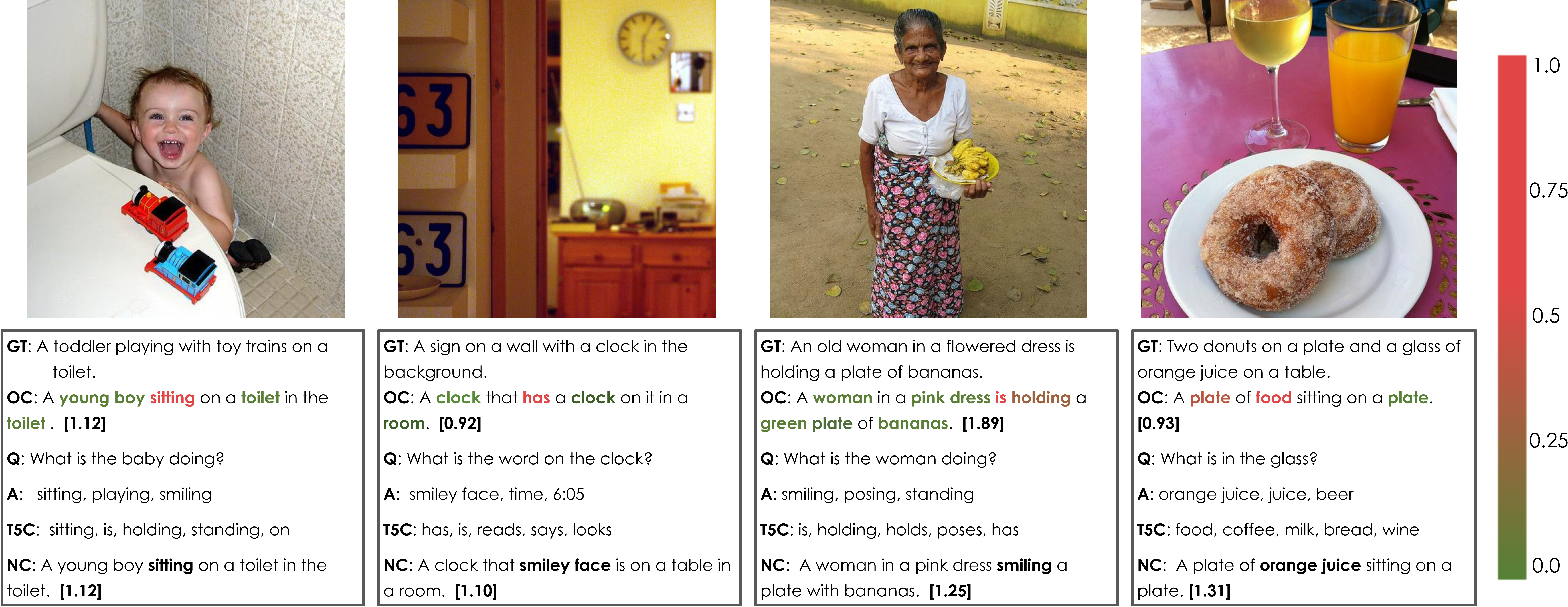}
\caption{Failure modes of our model. From left to right these images highlight the failures of: the decision maker, the question generator and VQA teacher, the captioner rolling out the rest of the sentence after receiving the answer, using auto-eval metrics as reward.}\label{fig:failcases}
\end{figure*}

Fig. \ref{fig:failcases} shows failure modes of our model. In the first image, the decision maker chooses a bad time to ask a question, and the agent gains no new information. In the second image, the question generator produces a bad question. The VQA teacher gives a nonsensical answer and the final caption is bad. The auto-eval metrics give the new caption a higher score than the original even though both captions are bad and it's unclear which one is better. In the third image, the captioning module is not able to utilize the answer from the expert. It ignores the answer in rolling out the rest of the sentence. In the last image, the agent is rewarded for identifying the orange juice in the image. However, the final sentence doesn't make grammatical sense. This is a limitation of using auto-eval metrics as the reward.

\end{document}